\documentclass{article}

    \usepackage[preprint]{neurips_2024}

\usepackage[utf8]{inputenc} 
\usepackage[T1]{fontenc}    
\usepackage{hyperref}       
\usepackage{url}            
\usepackage{booktabs}       
\usepackage{amsfonts}       
\usepackage{nicefrac}       
\usepackage{microtype}      
\usepackage{xcolor}         

\usepackage{multirow}
\usepackage{graphicx}
\usepackage{ulem}
\usepackage{caption}
\usepackage{subcaption}
\usepackage{amssymb}
\usepackage{booktabs}
\usepackage{multibib}
\usepackage{colortbl}
\usepackage{makecell}
\usepackage{wrapfig}
\usepackage{color}
\usepackage{float}
\usepackage{inconsolata}
\usepackage{amsmath}

\usepackage{enumitem}
\setlist{leftmargin=1em}

\usepackage[listings]{tcolorbox}
\tcbuselibrary{listings,theorems}
\newtcolorbox{mybox}{colback=white!5!white,colframe=black!75!black, left=.05in, right=.05in}

\definecolor{bluex}{rgb}{0.27, 0.42, 0.81}
\definecolor{purplex}{HTML}{9564bf}
\definecolor{red3}{HTML}{C52A20}
\definecolor{red2}{HTML}{B36A6F}
\definecolor{red1}{HTML}{FFb5b5}
\definecolor{purple}{HTML}{B36A6F}
\definecolor{darkyellow}{HTML}{D5BA82}
\definecolor{blue1}{HTML}{508AB2}
\definecolor{blue2}{HTML}{C4E4E3}
\definecolor{green1}{HTML}{A1D0C7}
\definecolor{green2}{HTML}{BFF6BA}
\definecolor{green3}{HTML}{028100}
\definecolor{teal}{HTML}{508AB2}
\definecolor{purple1}{HTML}{8d3a94}

\newtcbtheorem[]{exmp}{Example}%
{colback=purple1!5,colframe=purple1!80,fonttitle=\bfseries, left=.02in, right=.02in,bottom=.02in, top=.02in}{exmp}

\title{Make Your LLM Fully Utilize the Context}

\author{Shengnan An\thanks{\, Work done during the internship at Microsoft.}\hspace{0.4mm} $^{\diamondsuit,\clubsuit}$,\, Zexiong Ma$^{*\heartsuit,\clubsuit}$,\, Zeqi Lin\thanks{\, Corresponding authors.}\hspace{0.4mm} $^{\clubsuit}$,\\\textbf{Nanning Zheng}$^{\dagger\diamondsuit}$,\, \textbf{Jian-Guang Lou}$^{\clubsuit}$
\vspace{1mm}\\
  $^{\diamondsuit}$IAIR, Xi'an Jiaotong University,\,\,
  $^{\clubsuit}$Microsoft,\, 
  $^{\heartsuit}$Peking University\vspace{1mm}\\
  $^{\diamondsuit}$\texttt{\{an1006634493@stu, nnzheng@mail\}.xjtu.edu.cn}, \\
  $^{\heartsuit}$\texttt{mazexiong@stu.pku.edu.cn}, $^{\clubsuit}$\texttt{\{Zeqi.Lin, jlou\}@microsoft.com}
}

\begin{document}

\maketitle

\begin{abstract}

While many contemporary large language models (LLMs) can process lengthy input, they still struggle to fully utilize information within the long context, known as the \textit{lost-in-the-middle} challenge.
We hypothesize that it stems from insufficient explicit supervision during the long-context training, which fails to emphasize that any position in a long context can hold crucial information.
Based on this intuition, our study presents \textbf{\textsc{in}formation-\textsc{in}tensive (\textsc{In2}) training}, a purely data-driven solution to overcome lost-in-the-middle.
Specifically, \textsc{In2} training leverages a synthesized long-context question-answer dataset, where the answer requires (1) \textbf{fine-grained information awareness} on a short segment ($\sim$128 tokens) within a synthesized long context (4K$-$32K tokens), and (2) the \textbf{integration and reasoning} of information from two or more short segments.
Through applying this information-intensive training on Mistral-7B, we present \textbf{\textsc{FilM-7B}} (\textbf{\textsc{FIL}}l-in-the-\textbf{\textsc{M}}iddle).
To thoroughly assess the ability of \textsc{FilM-7B} for utilizing long contexts, we design three probing tasks that encompass various context styles (document, code, and structured-data context) and information retrieval patterns (forward, backward, and bi-directional retrieval).
The probing results demonstrate that \textsc{FilM-7B} can robustly retrieve information from different positions in its 32K context window.
Beyond these probing tasks, \textsc{FilM-7B} significantly improves the performance on real-world long-context tasks (e.g., 23.5$\rightarrow$26.9 F1 score on NarrativeQA), while maintaining a comparable performance on short-context tasks (e.g., 59.3$\rightarrow$59.2 accuracy on MMLU).
Github Link: \href{https://github.com/microsoft/FILM/tree/main}{github.com/microsoft/FILM}.

\begin{figure*}[h]
    \centering
    \includegraphics[width=.99\textwidth]{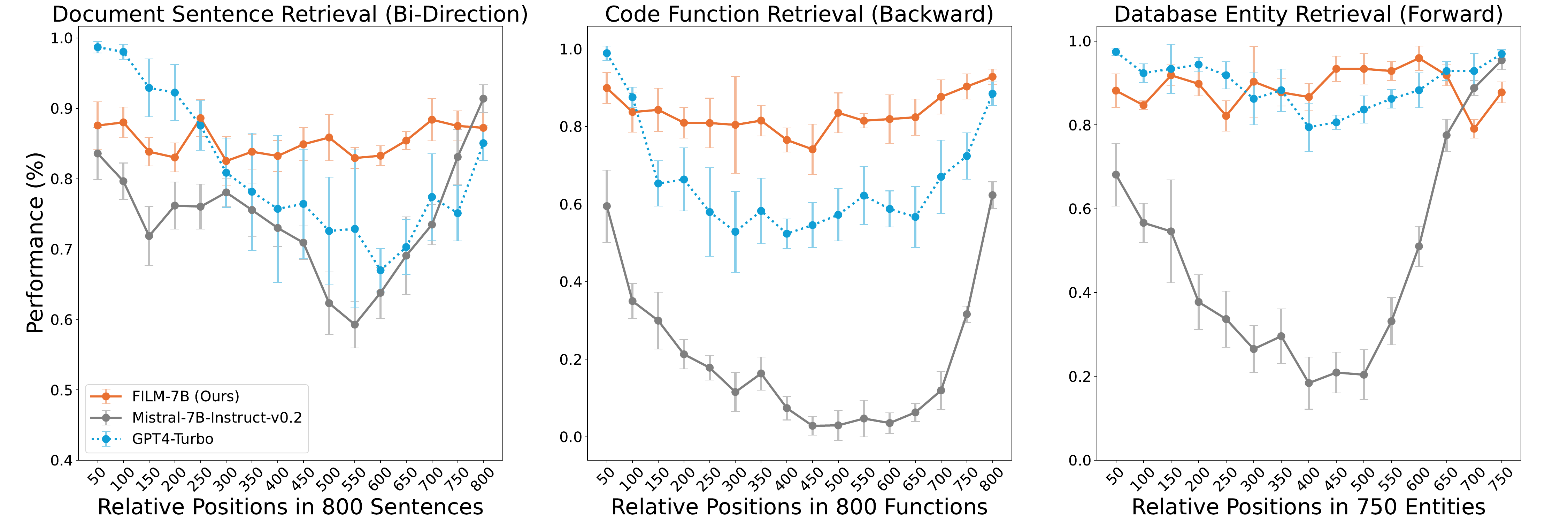}
    \caption{
    Performance of \textcolor[HTML]{E97132}{\textsc{FilM-7B}}, \textcolor[HTML]{7F7F7F}{Mistral-7B-Instruct-v0.2}, and \textcolor[HTML]{0F9ED5}{GPT4-Turbo} on our three probing tasks. \textsc{FilM-7B} significantly overcomes the problem of information loss in the middle of the context.
    }
    \label{fig:first}
\end{figure*}



\end{abstract}

\section{Introduction}

\begin{center}
    \fcolorbox{white}{white}{\parbox{.9\linewidth}{
    \centerline{\textit{To a great mind, nothing is little.}}
    \rightline{\textit{---Arthur Conan Doyle}}
}}
\end{center}

Long-context large language models (LLMs) have recently received significant attention within the open-source community~\citep{jiang2023mistral, du2022glm, li2023long, shi2023context, mosaicml2023introducing, together2023, chen2023extending, song2023hierarchical, liu2023scaling, peng2023yarn, chen2023longlora, xiong2023effective, tworkowski2024focused, ai2024yi, ding2024longrope, mohtashami2024random, fu2024data, cai2024internlm2, bai2024longalign, lv2024longwanjuan}.
The training context windows of many contemporary LLMs have been expanded to tens of thousands of tokens, thereby enabling these models to process extensive context as input.
This extended training context window can enhance many real-world downstream tasks such as long-context question answering~\citep{kovcisky2018narrativeqa, dasigi2021dataset, bai2023longbench} and summarization~\citep{fabbri2019multi, huang2021efficient, zhong2021qmsum}.


However, recent studies have revealed that these long-context LLMs struggle to effectively and robustly utilize all the information provided in the context, known as the \textit{lost-in-the-middle} challenge~\citep{liu2024lost, xu2023retrieval}.
It implies that while the LLM can comprehend the information at the beginning and end of the long context, it often overlooks the information in the middle.
This challenge could significantly hinder the development of long-context LLMs, as they even often fail to pass simple probing tasks such as Needle-in-the-Haystack and passkey retrieval~\citep{mohtashami2024random}.
Consequently, a pressing research question arises:
\textit{how can we make long-context LLMs fully utilize the information in the long context?}


We hypothesize that the root cause of lost-in-the-middle stems from the unintentional bias hidden in the general training data.
In auto-regressive pre-training, the loss on predicting the next token is more likely to be influenced by a few nearby pre-tokens rather than long-distance tokens~\citep{sharan2018prediction, sun2021long}.
For supervised fine-tuning and alignment, the system message, which strongly influences the generation of the response, is typically presented at the beginning of the context~\citep{touvron2023llama, cai2024internlm2}.
As a result, the general training process may inadvertently introduce a position bias, suggesting that important information is always located at the beginning and end of the context.


Based on this hypothesis, our work introduces \textbf{\textsc{in}formation-\textsc{in}tensive (\textsc{In2}) training} to explicitly teach the model that \textbf{the crucial information can be intensively present throughout the context}, not just at the beginning and end.
\textsc{In2} training is a purely data-driven solution that utilizes a synthesized long-context question-answer dataset.
The long context (ranging from 4K to 32K tokens) is concatenated from many short segments ($\sim$128 tokens), and the question-answer (QA) pairs ask for the information contained in one or more segments which are \textit{randomly} placed in the long context.
Specifically, we generate two types of questions, requiring (1) \textbf{fine-grained information awareness} on exactly one short segment, and (2) the \textbf{integration and reasoning of information} from two or more segments.
These QA pairs are generated by prompting GPT-4-Turbo~\citep{openai2023gpt4} with the designed instructions and the raw segments.


By applying this information-intensive training on Mistral-7B~\citep{jiang2023mistral}, we present \textbf{\textsc{FilM-7B}} (\textbf{\textsc{FIL}}l-in-the-\textbf{\textsc{M}}iddle).
To thoroughly assess the long-context information awareness of \textsc{FilM-7B}, we design three probing tasks encompassing various context styles (document, code, and structured-data context) and information retrieval patterns (forward, backward, and bi-directional retrieval).
The probing results (Figure~\ref{fig:first}) demonstrate that \textsc{In2} training significantly overcomes the lost-in-the-middle problem for the backbone model.
Moreover, it can enhance the open-source model to achieve comparable or even more robust performance compared with proprietary LLMs such as GPT-4-Turbo.


Beyond these probing tasks, the performance of \textsc{FilM-7B} on real-world long-context tasks also exhibits significant improvements (e.g., 23.5$\rightarrow$26.9 F1 score on NarrativeQA~\citep{kovcisky2018narrativeqa}).
This demonstrates that training on synthesized long-context data can be generalized to real-world scenarios.
Moreover, \textsc{FilM-7B} maintains a comparable performance on short-context tasks compared with the vanilla backbone model (e.g., 59.3$\rightarrow$59.2 accuracy on MMLU~\citep{hendrycks2020measuring}).
This indicates that the short-context capability of \textsc{FilM-7B} is not compromised during training.


The main contents of this paper are organized as follows.
Section~\ref{sec:in2} introduces our \textsc{In2} training with details on the data construction and training process.
Section~\ref{sec:probing} introduces the design of our long-context probing tasks and the comparison with some existing probing tasks.
Section~\ref{sec:main_results} shows the experimental results on three probing tasks, nine real-world long-context tasks, and eight short-context tasks.
Section~\ref{sec:training_strategy} provides further insights for the long-context training strategies.
Section~\ref{sec:related} discusses the related work.


\begin{figure*}[t]
    \centering
    \includegraphics[width=.85\textwidth]{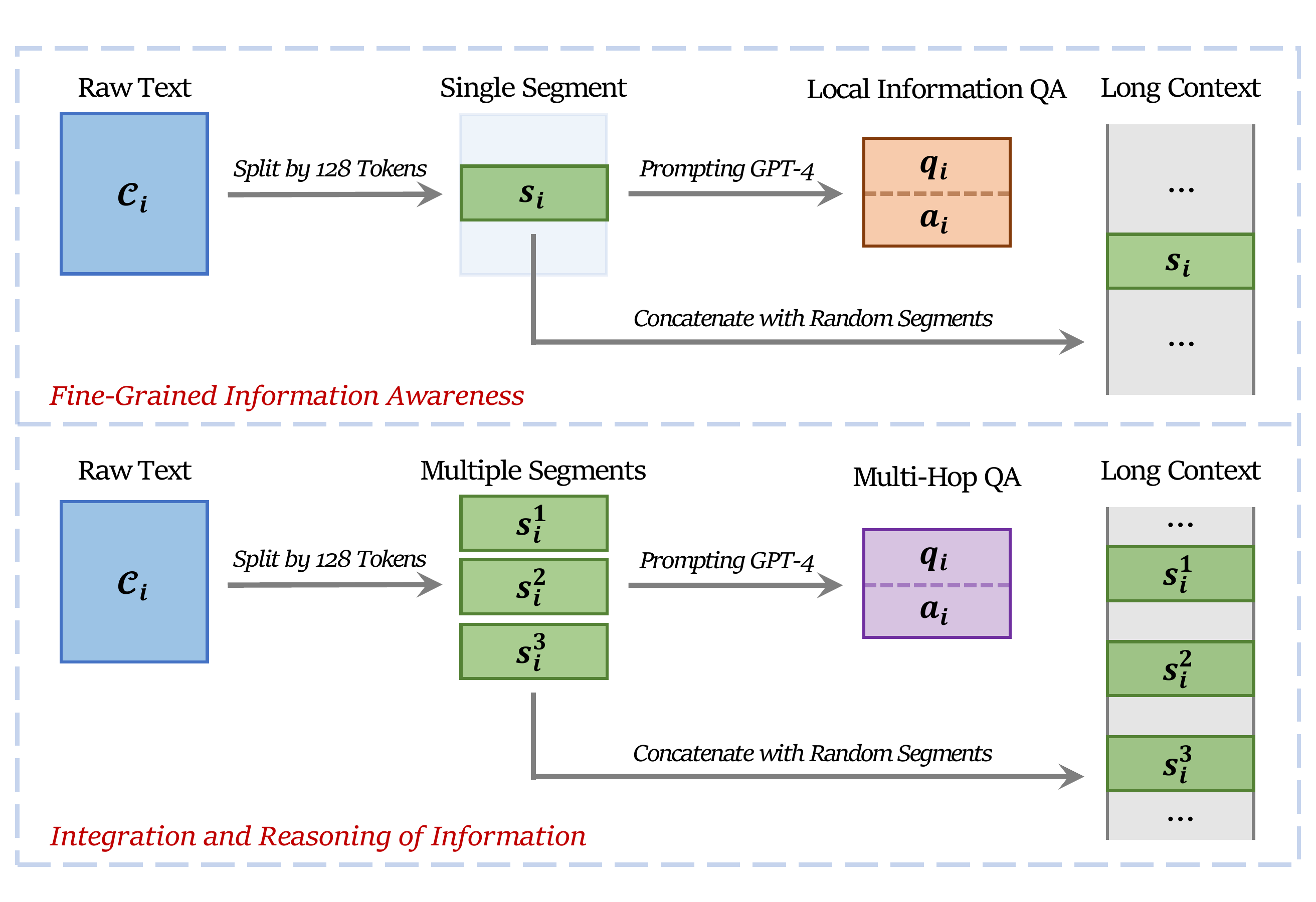}
    \caption{
    The data construction process for \textsc{In2} training, aimed at enhancing the fine-grained information awareness (upper), and the integration and reasoning of information (lower).
    }
    \label{fig:method}
\end{figure*}

\section{Information-Intensive Training}\label{sec:in2}

This section introduces the construction of the dataset for \textsc{In2} training and the detailed training process of our model \textsc{FilM-7B}.


\subsection{Training Data Construction}

\paragraph{Overview.}
The \textsc{In2} training aims to explicitly teach the model that any position in a long context can contain crucial information.
To achieve this goal, we construct a long-context question-answer training dataset $\mathbb{D}=\{\mathcal{L}_{i}, q_{i}, a_{i}\}$, where the answer $a_{i}$ to the question $q_{i}$ requires the information contained in some short segments that are randomly placed in the whole long context $\mathcal{L}_{i}$.

Figure~\ref{fig:method} illustrates an overview of the data construction process.
Specifically, the training data $\mathbb{D}$ is constructed based on a general natural language corpus $\mathbb{C}$.
Given a raw text $\mathcal{C}_{i}\in\mathbb{C}$, we first generate a question-answer pair $(q_{i}, a_{i})$ using a powerful LLM, then synthesize a long context $\mathcal{L}_{i}$ that includes the necessary information from $\mathcal{C}_{i}$ and other randomly sampled texts from $\mathbb{C}$.
We generate two types of question-answer pairs that require (1) the awareness of fine-grained information in the long context, and (2) the integration and reasoning of information appearing at different positions in the long context.
We take the \texttt{realnewslike} subset from the C4 corpus~\citep{raffel2020exploring} as $\mathbb{C}$, and take GPT-4-Turbo~\citep{openai2023gpt4} as the LLM to generate QA pairs.


\paragraph{Fine-grained information awareness.}
We consider a 128-token segment as the minimum information unit of the context\footnote{The raw texts in \texttt{realnewslike} have an average length of $\sim$600 tokens with the Mistral tokenizer.}.
Given a raw text $\mathcal{C}_{i}$, we first randomly extract a 128-token segment $s_{i}$ from it, then generate the $q_{i}$, $a_{i}$ and $\mathcal{L}_{i}$ accordingly,
\begin{equation}\label{equ:fine}
    (q_{i}, a_{i})\sim\mathrm{Prompting}(s_{i}, I_{f};\mathrm{LLM}),\quad \mathcal{L}_{i}=\oplus\{\mathrm{Shuffle}(s_{i}, [r_{j}])\},
\end{equation}
where $(q_{i}, a_{i})$ is sampled by prompting the powerful LLM with the segment $s_{i}$ and the instruction $I_{f}$, $\oplus\{\cdot\}$ represents the concatenation of the contained segments, and $[r_{j}]$ are randomly sampled from 128-token segments in $\mathbb{C}$.
Note that $I_{f}$ instructs the LLM to make the question-answer pair highly specific to the information provided in $s_{i}$.

\paragraph{Integration and reasoning of information.}
Beyond utilizing each single segment, we consider to generate question-answer pairs for information contained in two or more segments.
Following the setting of the minimum information unit above,
we split a full text $\mathcal{C}_{i}$ into a set of 128-token segments $[s_{i}]$, then generate the $q_{i}$, $a_{i}$ and $\mathcal{L}_{i}$ accordingly,
\begin{equation}\label{equ:reason}
    (q_{i}, a_{i})\sim\mathrm{Prompting}([s_{i}], I_{r};\mathrm{LLM}),\quad \mathcal{L}_{i}=\oplus\{\mathrm{Shuffle}([s_{i}], [r_{j}])\},
\end{equation}
where $I_{r}$ instructs the LLM to generate a multi-hop question-answer pair that requires the information within at least two segments in $[s_{i}]$.
All segments in $[s_{i}]$ and $[r_{j}]$ are jointly shuffled, so the required segments may appear far apart in the context.

\paragraph{Context length balance and data mixture.}
To prevent length bias during \textsc{In2} training, we ensure the length of the long context $\mathcal{L}_{i}$ is evenly distributed from 4K to 32K tokens.
Such a length balance strategy can be implemented with reject sampling on $[r_{j}]$, according to Equation~\ref{equ:fine} and \ref{equ:reason}.
To alleviate catastrophic forgetting on short-context capabilities, we retain $\sim$10\% question-answer pairs with the original texts $\mathcal{C}_{i}$ instead of converting them into a longer context, and add some general instruction-tuning data from the OpenOrca~\citep{OpenOrca} dataset.

Overall, our dataset for \textsc{In2} training contains 1.1M long-context data for the fine-grained information awareness ($\sim$63\%), 300K long-context data for the integration and reasoning of information ($\sim$17\%), 150K short-context question-answer data ($\sim$9\%), and 200K general instruction-tuning data ($\sim$11\%).
Appendix~\ref{sec:ap_prompts} contains the handcraft instructions for data generation.
Appendix~\ref{sec:ap_examples} illustrates some examples of our constructed long-context QA data.
Appendix~\ref{sec:ap_filter} describes the filtering strategy to avoid data contamination for evaluation.

\begin{figure*}[t]
    \centering
    \includegraphics[width=.99\textwidth]{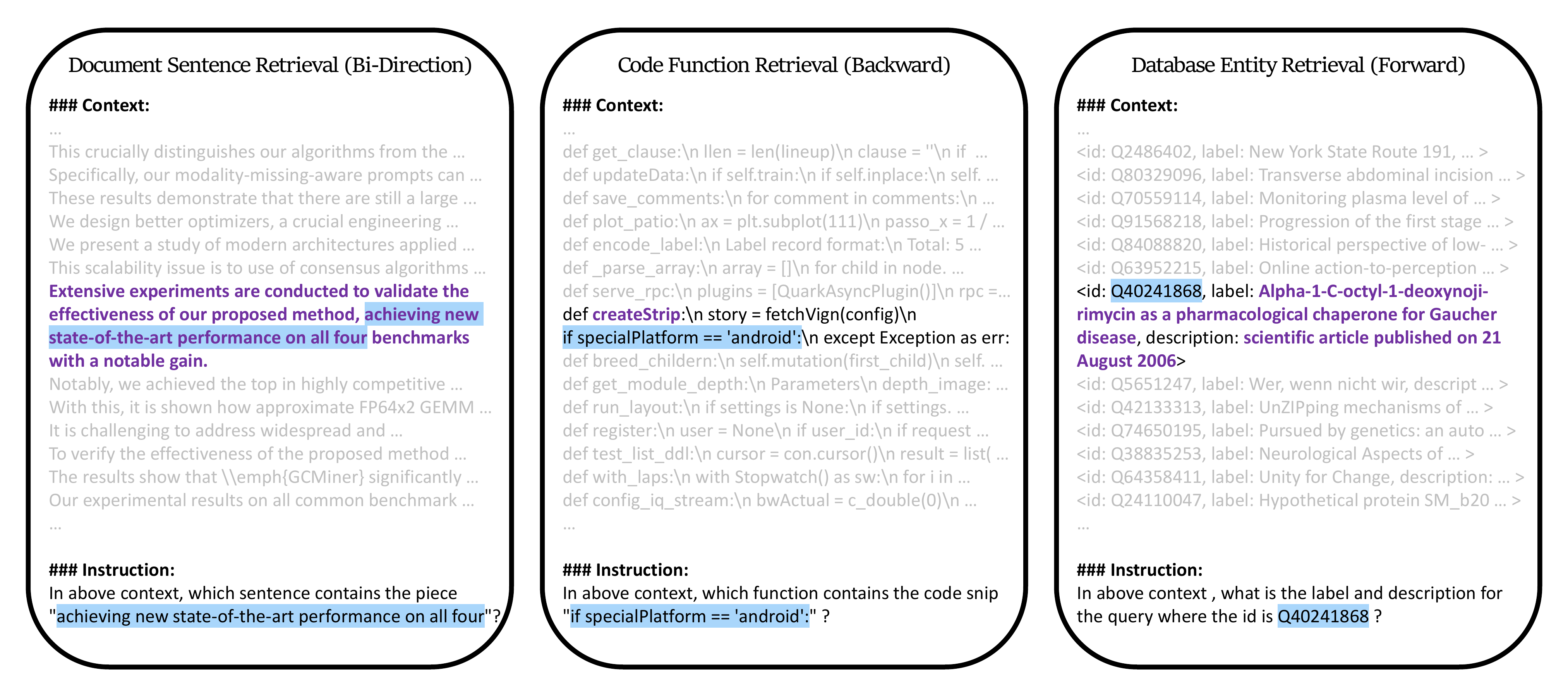}
    \caption{
    Three tasks in \textsc{VaL} Probing. The retrieval patterns are determined by the relative positions between the \colorbox[HTML]{A9D6FB}{retrieval keywords} and the \textcolor[HTML]{7030A0}{\textbf{information to be retrieved}}.
    }
    \label{fig:probing}
\end{figure*}

\subsection{Training Details}

Using the training data constructed above, we further fine-tune the Mistral-7B-Instruct-v0.2\footnote{\href{https://huggingface.co/mistralai/Mistral-7B-Instruct-v0.2}{https://huggingface.co/mistralai/Mistral-7B-Instruct-v0.2}.}~\citep{jiang2023mistral} to get our \textsc{FilM-7B} (\textbf{\textsc{FIL}}l-in-the-\textbf{\textsc{M}}iddle).
We perform \textsc{In2} training in the instruction-tuning paradigm:
the long contexts and questions are used as instructions, and the loss on the answer parts are used to update the model.
Appendix~\ref{sec:ap_prompts} contains the system template used for formatting the training data.
For hyper-parameters, we set the global batch size as 128 and conduct one-epoch training with $\sim$14K training steps.
We use the cosine learning rate decay with a 1e-6 maximum learning rate and 3\% warm-up steps.
The training process is conducted on 16 nodes of 8x80G A100 GPUs with the full sharding strategy and cpu offload strategy implemented by pytorch FSDP~\citep{zhao2023pytorch}.
The entire training process consumes $\sim$300 GPU days.

\begin{figure*}[t]
	\centering
    \begin{subfigure}[t]{0.98\textwidth}
         \centering
         \includegraphics[width=\textwidth]{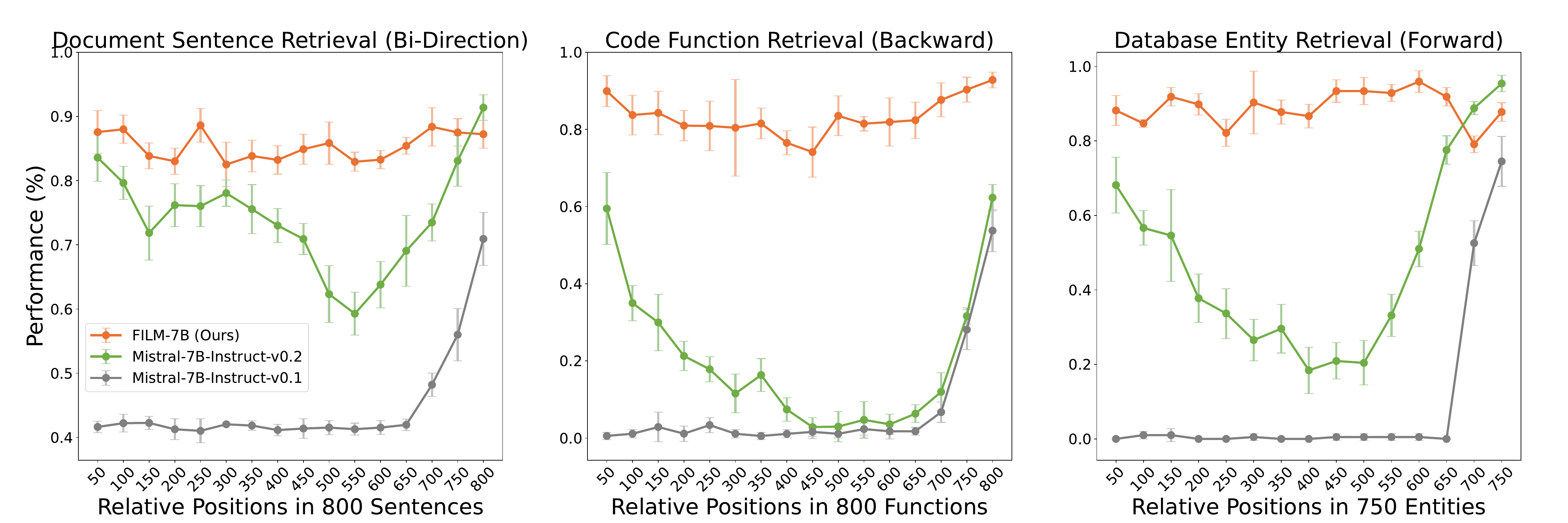}
         \caption{Performance of \textcolor[HTML]{E97132}{\textsc{FilM-7B}}, \textcolor[HTML]{7F7F7F}{Mistral-7B-Instruct-v0.1}, and \textcolor[HTML]{70AD47}{Mistral-7B-Instruct-v0.2}.}
         \label{fig:probing_mistral}
     \end{subfigure}\\
     \begin{subfigure}[t]{0.98\textwidth}
         \centering
         \includegraphics[width=\textwidth]{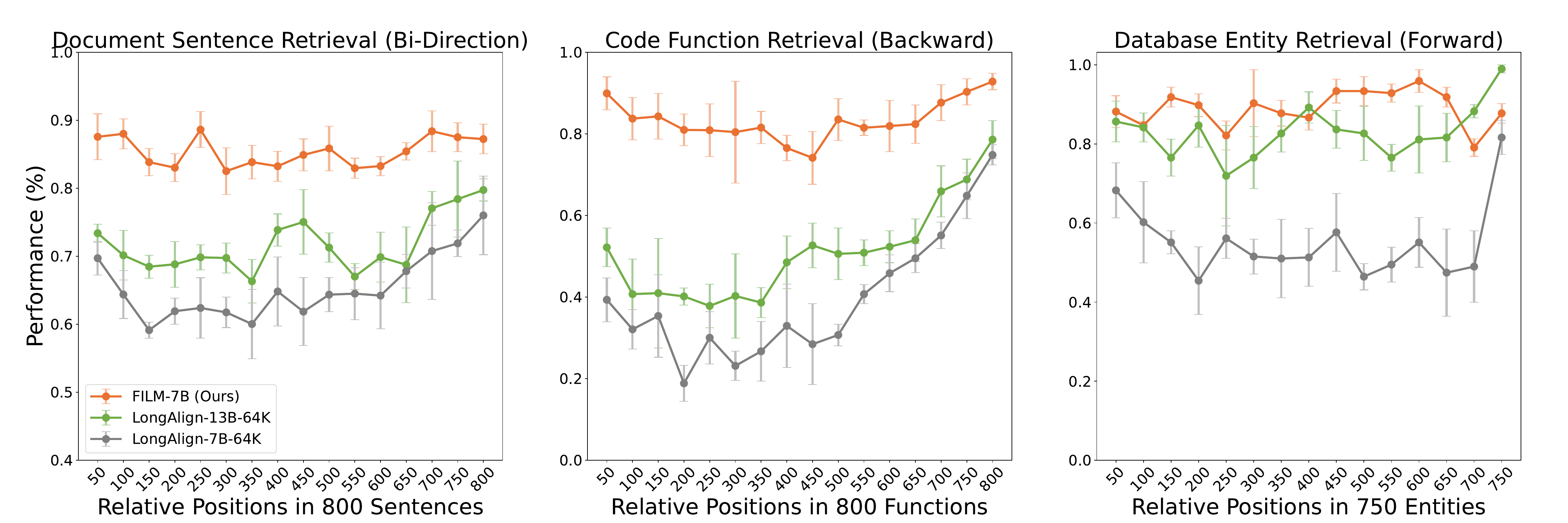}
         \caption{Performance of \textcolor[HTML]{E97132}{\textsc{FilM-7B}}, \textcolor[HTML]{7F7F7F}{LongAlign-7B-64K}, and \textcolor[HTML]{70AD47}{LongAlign-13B-64K}.}
         \label{fig:probing_longalign}
     \end{subfigure}\\
     \begin{subfigure}[t]{0.98\textwidth}
         \centering
         \includegraphics[width=\textwidth]{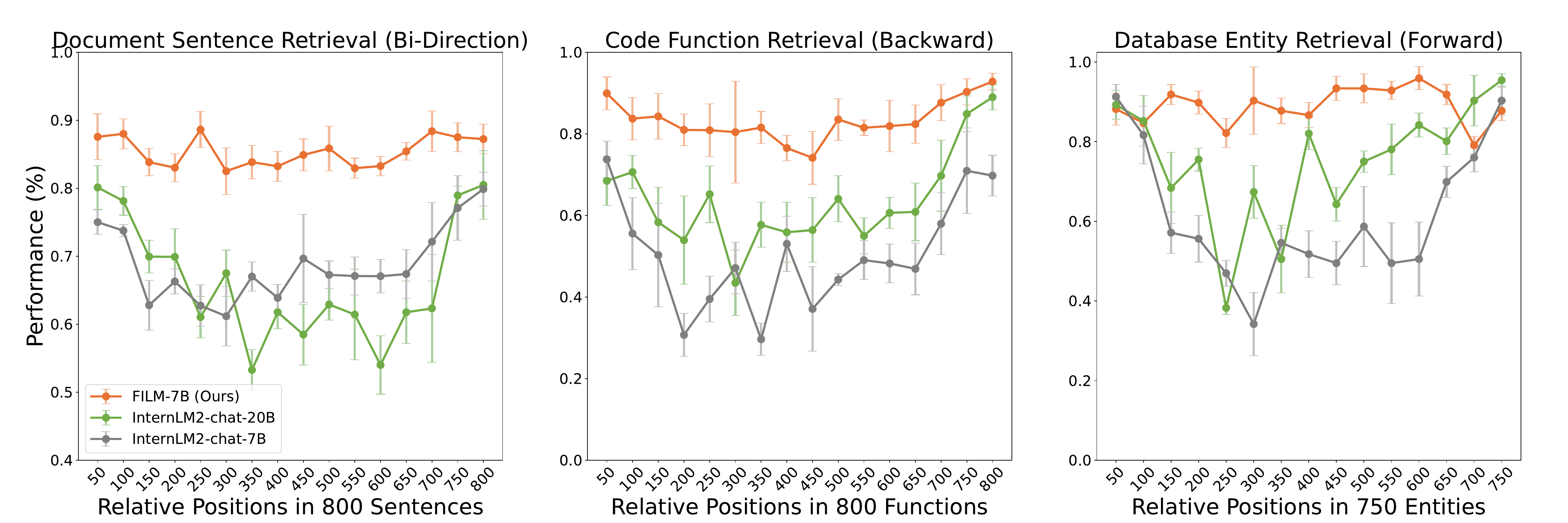}
         \caption{Performance of \textcolor[HTML]{E97132}{\textsc{FilM-7B}}, \textcolor[HTML]{7F7F7F}{InternLM2-chat-7B}, and \textcolor[HTML]{70AD47}{InternLM2-chat-20B}.}
         \label{fig:probing_internlm}
     \end{subfigure}
	\caption{Performance of \textsc{FilM-7B} on \textsc{VaL} Probing and the comparisons with (a) Mistral, (b) LongAlign, and (c) InternLM2. The X-axis is the relative position in the context ($\sim$32K tokens).}
    \label{fig:probing_results}
\end{figure*}

\section{Long-Context Probing}\label{sec:probing}

In this section, we first show the preliminary evaluation of \textsc{FilM-7B} on the Needle-in-the-Haystack and discuss about the inadequacies of this probing task.
Subsequently, to comprehensively evaluate the long-context information awareness of \textsc{FilM-7B}, we introduce \textbf{\textsc{Va}rious \textsc{L}ong-context (\textsc{VaL}) Probing}. This includes three tasks that cover various context styles (document, code, and structured-data context) and information retrieval patterns (forward, backward, and bi-directional retrieval).

\subsection{Near-Perfect Performance on Needle-in-the-Haystack: Are We There Yet?}

The Needle-in-the-Haystack\footnote{\href{https://github.com/gkamradt/LLMTest_NeedleInAHaystack}{https://github.com/gkamradt/LLMTest\_NeedleInAHaystack}.} task is widely used to assess how robustly a model utilizes information positioned in the long context.
It reveals that even some powerful proprietary LLMs, such as GPT-4 and Claude 2.1~\citep{claude2}, struggle to fully exploit the information within the long context.


We use the Needle-in-the-Haystack task to preliminarily evaluate the long-context capability of \textsc{FilM-7B}.
Appendix~\ref{sec:ap_needle} demonstrates that \textsc{FilM-7B} has achieved near-perfect performance on this task.
This result is not surprising as recent open-source LLMs, such as LongAlign~\citep{bai2024longalign} and InternLM2~\citep{cai2024internlm2}, have also shown near-perfect performance on this task.


However, we argue that the near-perfect performance on Needle-in-the-Haystack may overestimate the long-context capabilities of LLMs, based on the following two considerations:
\begin{itemize}
\item Needle-in-the-Haystack employs a document-style context, which LLMs could be quite familiar with due to the pre-training on natural language corpora.
\item The \textbf{forward retrieval} pattern in Needle-in-the-Haystack may simplify the difficulty of information seeking in the long context.
\end{itemize}
The ``forward retrieval'' means that the information being retrieved directly follows the retrieval keyword in a long context.
For example, the default question used in Needle-in-the-Haystack is "What is the best thing to do in San Francisco?" and the answer is contained in "The best thing to do in San Francisco is eat a sandwich and sit in Dolores Park on a sunny day."
The retrieved information "eat a sandwich and ..." just follows the retrieval keywords "best thing to do in San Francisco".
According to the mechanism of induction head~\citep{olsson2022incontext}, such a following-up copying is an easily learned pattern for LLMs, thus less challenging for evaluating long context utilization.

Given these considerations, we suggest that performances on Needle-in-the-Haystack may not adequately reflect the long-context capabilities of LLMs.
Therefore, we propose \textsc{VaL} Probing for a more comprehensive evaluation involving various context styles and retrieval patterns.



\begin{table*}[t]
\renewcommand\arraystretch{1.2}
\caption{Quantified performances of various models on \textsc{VaL} Probing.
}
\label{tab:probing_results}
\centering
\resizebox{.9\linewidth}{!}{
\begin{tabular}{@{}lcccccccc@{}}
\toprule
\multicolumn{1}{c}{\multirow{2}{*}{Model}} & \multicolumn{2}{c}{Document} & \multicolumn{2}{c}{Code} & \multicolumn{2}{c}{Database} & \multicolumn{2}{c}{All} \\
\multicolumn{1}{c}{} & Avg & Gap$\downarrow$ & Avg & Gap$\downarrow$ & Avg & Gap$\downarrow$ & Avg & Gap$\downarrow$ \\ \midrule
Mistral-7B-Instruct-v0.1~\citep{jiang2023mistral} & 44.8 & 29.9 & 6.8 & 53.2 & 8.8 & 74.5 & 20.1 & 52.5 \\
Mistral-7B-Instruct-v0.2~\citep{jiang2023mistral} & 74.2 & 32.1 & 20.3 & 59.5 & 47.5 & 77.0 & 47.3 & 56.2 \\
LongAlign-7B-64K~\citep{bai2024longalign} & 65.3 & 16.9 & 39.3 & 56.0 & 55.0 & 36.2 & 53.2 & 36.4 \\
LongAlign-13B-64K~\citep{bai2024longalign} & 71.7 & 13.4 & 50.8 & 40.8 & 82.9 & 27.0 & 68.5 & 27.1 \\
InternLM2-chat-7B~\citep{cai2024internlm2} & 68.8 & 18.7 & 50.2 & 44.1 & 61.2 & 57.1 & 60.1 & 40.0 \\
InternLM2-chat-20B~\citep{cai2024internlm2} & 66.4 & 27.2 & 63.4 & 45.5 & 74.9 & 57.2 & 68.2 & 43.3 \\
GPT-4-Turbo~\citep{openai2023gpt4} & 81.3 & 31.7 & 66.1 & 46.5 & \textbf{89.6} & 18.0 & 79.0 & 32.1 \\
\cellcolor{gray!25}\textsc{FilM-7B} (ours) & \cellcolor{gray!25}\textbf{85.4} & \cellcolor{gray!25}\textbf{6.1} & \cellcolor{gray!25}\textbf{83.3} & \cellcolor{gray!25}\textbf{18.7} & \cellcolor{gray!25}\textbf{89.0} & \cellcolor{gray!25}\textbf{16.8} & \cellcolor{gray!25}\textbf{85.9} & \cellcolor{gray!25}\textbf{13.9} \\ \bottomrule
\end{tabular}
}
\end{table*}

\subsection{\textsc{VaL} Probing}

Our retrieval-based \textsc{VaL} Probing considers three context styles (document, code, and structured-data context) and three retrieval patterns (forward, backward, and bi-directional retrieval).
Each context in \textsc{VaL} Probing contains $\sim$32K tokens, and each task contains $\sim$3K examples. Figure~\ref{fig:probing} briefly illustrates the contexts and retrieval instructions in \textsc{VaL} Probing.


\paragraph{Document Sentence Retrieval (Bi-Direction).}
The contexts consist of numerous natural language sentences, and the instruction aims to retrieve a single sentence containing a given piece.
The sentences are sampled from the abstracts of papers on arXiv\footnote{\href{https://info.arxiv.org/help/api/basics.html}{https://info.arxiv.org/help/api/basics.html}.}.
This task follows the bi-directional retrieval pattern, as the expected retrieval results contain words both before and after the given piece in the context.
The evaluation metric is the word-level recall score.

\paragraph{Code Function Retrieval (Backward).}
The contexts consist of Python functions, and the instruction aims to retrieve the function name for a given line of code within the function definition.
The raw code functions are sampled from the StarCoder~\citep{li2023starcoder} dataset\footnote{\href{https://huggingface.co/datasets/bigcode/starcoderdata}{https://huggingface.co/datasets/bigcode/starcoderdata}.}. 
We randomly select three lines of definitions for each function.
This task follows the backward retrieval pattern, as the function name always precedes the definition.
The evaluation metric is the exact-match accuracy.

\paragraph{Database Entity Retrieval (Forward).}
The contexts contain lists of structured entities, each with three fields: ID, label, and description.
The query aims to retrieve the label and description for a given ID.
The entities are sampled from Wikidata~\footnote{\href{https://www.wikidata.org/wiki/Wikidata:Data\_access}{https://www.wikidata.org/wiki/Wikidata:Data\_access}.}. 
This task follows the forward retrieval pattern, as the label and description follow the ID.
We take a relaxed exact-match accuracy as the metric: a 1 score is given if either the label or the description is exactly matched in the response, otherwise a 0 score.

\begin{table*}[t]
\renewcommand\arraystretch{1.2}
\Huge
\caption{Performances of various models on real-world long-context tasks. Results of models with $^{*}$ are reported in~\citet{bai2023longbench} and~\citet{lv2024longwanjuan}.
}
\label{tab:real_world}
\centering
\resizebox{.99\linewidth}{!}{
\begin{tabular}{@{}lcccccccccc@{}}
\toprule
\multicolumn{1}{c}{Model} & NarrativeQA & Qasper & MultiFQA & HotpotQA & 2WikiMQA & MuSiQue & GovReport & QMSum & MultiNews & Avg \\ \midrule\midrule
\multicolumn{11}{c}{Close-Source} \\
GPT-4-Turbo~\citep{openai2023gpt4} & 33.0 & 50.7 & 52.7 & 68.5 & 64.3 & 49.1 & 33.9 & 25.4 & 24.9 & 44.7 \\
GPT-3.5-Turbo$^{*}$~\citep{gpt35turbo} & 23.6 & 43.3 & 52.3 & 51.6 & 37.7 & 26.9 & 29.5 & 23.4 & 26.7 & 35.0 \\ \midrule\midrule
\multicolumn{11}{c}{Open-Source} \\
LongChat-v1.5-7B-32K$^{*}$~\citep{li2023long} & 16.9 & 27.7 & 41.4 & 31.5 & 20.6 & 9.7 & 30.8 & 22.7 & 26.4 & 25.3 \\
ChatGLM2-6B-32K$^{*}$~\citep{du2022glm} & 21.1 & 31.5 & 46.2 & 25.3 & 20.8 & 9.8 & 32.4 & 24.0 & 26.5 & 26.4 \\
LongAlign-7B-64K~\citep{bai2024longalign} & 18.7 & 33.8 & 49.1 & 28.6 & 23.4 & 12.5 & 30.6 & 23.7 & 27.5 & 27.5 \\
Mistral-7B-Instruct-v0.1~\citep{jiang2023mistral} & 19.6 & 33.2 & 38.8 & 42.9 & 31.2 & 17.4 & 27.5 & 22.4 & 26.6 & 28.9 \\
Mistral-7B-Instruct-v0.2~\citep{jiang2023mistral} & 23.5 & 33.8 & 45.9 & 42.4 & 24.3 & 20.8 & 33.3 & 24.8 & 26.8 & 30.6 \\
Yi-6B-200K$^{*}$~\citep{ai2024yi} & 12.4 & 26.4 & 36.8 & 46.6 & 40.4 & 25.8 & 29.3 & 20.7 & 27.1 & 29.5 \\
ChatGLM3-6B-32K$^{*}$~\citep{du2022glm} & 9.2 & \textbf{43.1} & 50.9 & 55.3 & 43.7 & \textbf{38.9} & \textbf{36.0} & 24.7 & 27.4 & 36.6 \\
InternLM2-chat-7B~\citep{cai2024internlm2} & 24.4 & 35.4 & 50.2 & 52.4 & \textbf{48.2} & 30.5 & 33.6 & \textbf{25.3} & \textbf{29.0} & 36.5 \\
InternLM2-7B-LongWanjuan$^{*}$~\citep{lv2024longwanjuan} & \textbf{29.9} & 39.6 & 50.2 & 53.7 & 42.3 & 32.1 & 33.0 & \textbf{25.5} & 27.8 & 37.1 \\
\cellcolor{gray!25}\textsc{FilM-7B}   (ours) & \cellcolor{gray!25}26.9 & \cellcolor{gray!25}\textbf{42.2} & \cellcolor{gray!25}\textbf{56.0} & \cellcolor{gray!25}\textbf{62.1} & \cellcolor{gray!25}\textbf{47.0} & \cellcolor{gray!25}\textbf{39.0} & \cellcolor{gray!25}33.8 & \cellcolor{gray!25}\textbf{25.1} & \cellcolor{gray!25}26.9 & \cellcolor{gray!25}\textbf{39.9} \\ \bottomrule
\end{tabular}
}
\end{table*}

\section{Experiments and Analysis}\label{sec:experiments}

We assess the long-context capability of \textsc{FilM-7B} on both probing tasks and real-world long-context tasks. Moreover, we investigate if the performance in short-context scenarios is affected.


\subsection{Experimental Setup}

\paragraph{Models.}
We mainly compare \textsc{FilM-7B} with long-context open-source models that have been trained with $\geq$32K context windows, including the Mistral~\citep{jiang2023mistral}, LongChat~\citep{li2023long}, ChatGLM~\citep{du2022glm}, LongAlign~\citep{bai2024longalign}, LongWanjuan~\citep{lv2024longwanjuan}, Yi~\citep{ai2024yi} and InternLM2~\citep{cai2024internlm2}.
We utilize the instruct/chat versions of these models as most of our evaluation tasks are under the zero-shot instruction-following paradigm.
We also draw comparisons with popular proprietary LLMs such as GPT-3.5-Turbo~\citep{gpt35turbo} and GPT-4-Turbo~\citep{openai2023gpt4}.
All models and tasks employ greedy decoding.
For probing tasks, we primarily compare \textsc{FilM-7B} with LongAlign and InternLM2 series, as these models have shown near-perfect performances on Needle-in-the-Haystack.

\paragraph{Real-world long-context tasks.}
We take 9 tasks from the LongBench~\citep{bai2023longbench} collection to evaluate the long-context capability on real-world scenarios.
These tasks encompass long-document question answering (NarrativeQA~\citep{kovcisky2018narrativeqa}, Qasper~\citep{dasigi2021dataset} and MultiFieldQA (MultiFQA)~\citep{bai2023longbench}, multi-document multi-hop reasoning (HotpotQA~\citep{yang2018hotpotqa}, 2WikiMultihopQA (2WikiMQA)~\citep{ho2020constructing} and MuSiQue~\citep{trivedi2022musique}), and long-context summarization (GovReport~\citep{huang2021efficient}, QMSum~\citep{zhong2021qmsum} and MultiNews~\citep{fabbri2019multi}).
We employ the middle truncation strategy in LongBench to limit the input within 32K tokens.
We report ROUGE-L~\citep{lin2004rouge} for summarization tasks and F1 scores for other tasks.
The evaluation metrics are computed using the official evaluation scripts~\footnote{\href{https://github.com/THUDM/LongBench}{https://github.com/THUDM/LongBench}.}.

\paragraph{Short-context tasks.}
We select 8 short-context tasks commonly used for evaluating the general capabilities of models.
These include MMLU~\citep{hendrycks2020measuring}, BoolQ~\citep{clark2019boolq}, RACE-High (RACE-H)~\citep{lai2017race}, CommonsenseQA (CSQA)~\citep{commonsenseqa2019}, ARC-Challenge (ARC-C)~\citep{clark2018think}, HellaSwag~\citep{zellers2019hellaswag}, GSM8K~\citep{cobbe2021training}, and MATH~\citep{hendrycks2021measuring}.
We use 5-shot for MMLU, 8-shot for GSM8K, 4-shot for MATH, and 0-shot for other tasks.
We utilize the \texttt{lm\_eval}\footnote{\href{https://github.com/EleutherAI/lm-evaluation-harness}{https://github.com/EleutherAI/lm-evaluation-harness}.} for the evaluations on MMLU, BoolQ, RACE-H, ARC-C and HellaSwag, and use the evaluation scripts from~\citet{an2024learning} for other tasks.

\begin{figure*}[t]
    \centering
    \includegraphics[width=.99\textwidth]{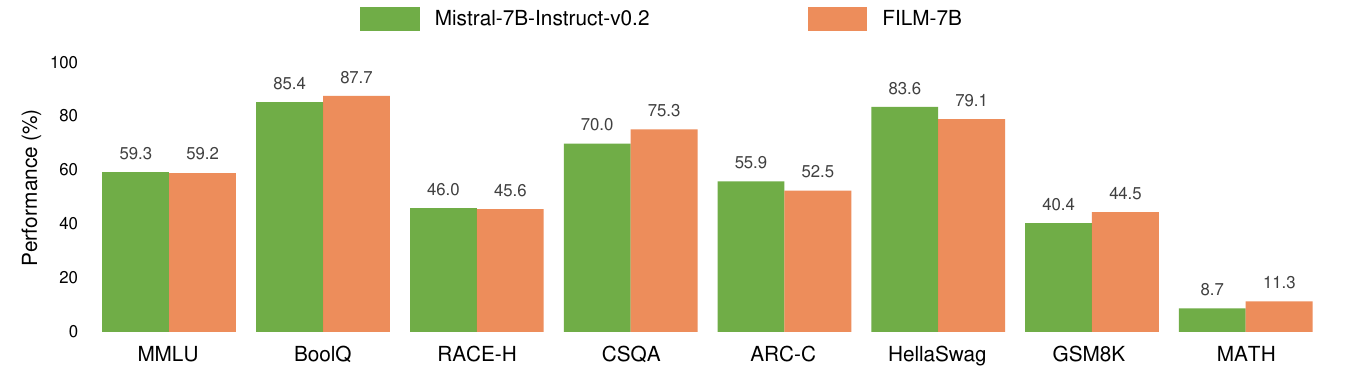}
    \caption{
    Performances of \textsc{FilM-7B} and the backbone model on short-context tasks.
    }
    \label{fig:short}
\end{figure*}

\subsection{Main Results}\label{sec:main_results}

\paragraph{\textsc{FilM-7B} significantly mitigates the lost-in-the-middle problem.}
Figure~\ref{fig:probing_mistral} presents the probing results for both \textsc{FilM-7B} and the backbone model, Mistral-7B-Instruct-v0.2.
In all three probing tasks within \textsc{Val} Probing, the vanilla Mistral model experiences substantial information loss at the middle positions in the long contexts.
In contrast, our \textsc{FilM-7B} model consistently exhibits robust performance across different positions within the whole context. 
This stark comparison illustrates that the lost-in-the-middle problem can be effectively addressed using our \textsc{In2} training.

\paragraph{\textsc{FilM-7B} achieves performance comparable to, or even outperforming, that of GPT-4-Turbo.}
Figure~\ref{fig:first} illustrates the comparison between \textbf{FILM-7B} and GPT-4-Turbo on our probing tasks.
Beyond a qualitative comparison between the performance curves of two models, we quantify the long-context performances on \textsc{VaL} Probing using two metrics:
\begin{itemize}
\item \textbf{Average score (Avg).} We compute the average performances across the entire context length, reflecting the overall long-context utilization.
\item \textbf{Min-max gap (Gap).} We calculate the differences between the maximum and minimum performances in Figure~\ref{fig:probing}. A smaller performance gap signifies greater robustness across different positions.
\end{itemize}
Table~\ref{tab:probing_results} presents the quantified performances on \textsc{VaL} Probing.
It reveals that \textsc{FilM-7B} has comparable performance with GPT-4-Turbo on the database probing task, and exhibits better robustness in document and code probing tasks.
These results indicate a great potential for the development of open-source long-context models to close the gap with proprietary models.


\paragraph{\textsc{VaL} Probing presents a more challenging test suite for long-context models.}
Figure~\ref{fig:probing_longalign} and~\ref{fig:probing_internlm} show the probing results of LongAlign and InternLM2, two state-of-the-art long-context models.
Despite their extended training context windows, these models still encounter the lost-in-the-middle problem.
This is particularly noteworthy given their near-perfect performance on the Needle-in-the-Haystack task. 
This comparison suggests that \textsc{VaL} Probing provides a more challenging evaluation for long-context models.

In particular, the results on document and database tasks in \textsc{VaL} Probing demonstrate clear comparisons with Needle-in-the-Haystack.
Compared to Needle-in-the-Haystack which uses forward retrieval on natural language context, the document task employs natural language context but with bi-directional retrieval, and the database task uses forward retrieval but with structured-data context.
These comparisons highlight that both context styles and retrieval patterns significantly contribute to the hardness of the probing tasks.



\begin{figure*}[t]
    \centering
    \includegraphics[width=.99\textwidth]{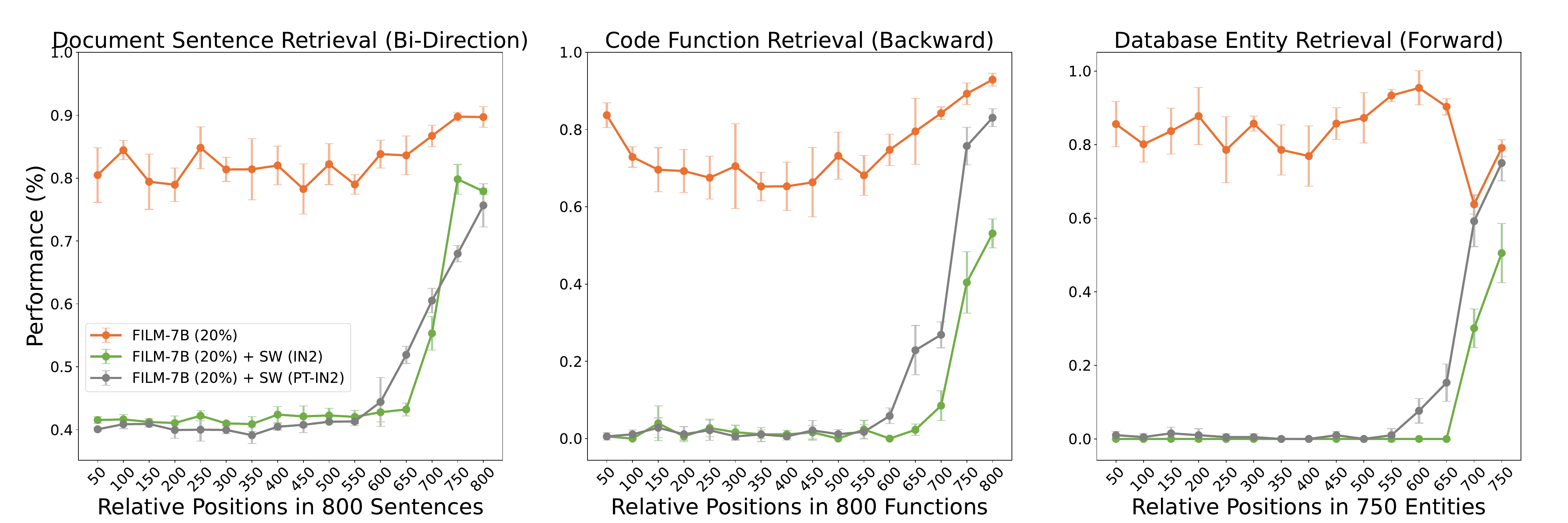}
    \caption{
    Performance of \textsc{FilM-7B} with a 4K sliding window (SW).
    PT-\textsc{In2}: apply the sliding window in both pre-training and \textsc{In2} training.
    \textsc{In2}: apply the sliding window only in \textsc{In2} training.
    }
    \label{fig:sw}
\end{figure*}

\begin{table*}[t]
\renewcommand\arraystretch{1.2}
\caption{Performance of \textsc{FilM-7B} with different RoPE base $\theta$ during \textsc{In2} training.
}
\label{tab:rope}
\centering
\resizebox{.85\linewidth}{!}{
\begin{tabular}{@{}cccccccccc@{}}
\toprule
\multirow{2}{*}{Model} & \multirow{2}{*}{RoPE Base $\theta$} & \multicolumn{2}{c}{Document} & \multicolumn{2}{c}{Code} & \multicolumn{2}{c}{Database} & \multicolumn{2}{c}{All} \\
 &  & Avg & Gap$\downarrow$ & Avg & Gap$\downarrow$ & Avg & Gap$\downarrow$ & Avg & Gap$\downarrow$ \\ \midrule
\multirow{4}{*}{\textsc{FILM-7B}   (20\%)} & $1.0\times10^6$  (default) & 82.9 & 11.5 & 74.5 & 27.7 & 83.5 & 31.6 & 80.3 & 23.6 \\
 & $2.0\times10^6$ & 83.9 & 9.3 & 79.8 & 27.1 & 87.7 & \textbf{13.2} & 83.8 & 16.5 \\
 & $1.0\times10^7$ & 83.7 & \textbf{7.6} & \textbf{81.7} & \textbf{18.4} & \textbf{89.4} & 16.8 & \textbf{84.9} & \textbf{14.3} \\
 & $1.0\times10^8$ & \textbf{84.6} & \textbf{6.6} & \textbf{81.4} & 22.3 & 87.7 & \textbf{13.2} & \textbf{84.6} & \textbf{14.0} \\ \bottomrule
\end{tabular}
}
\end{table*}

\paragraph{Training on synthesized long-context data effectively generalizes to real-world scenarios.}
Table~\ref{tab:real_world} contains the results on various real-world long-context tasks.
It shows that \textsc{FilM-7B} also significantly improves the performance of the backbone model in real-world long-context scenarios.
Moreover, it also achieves SOTA-level performances on these tasks among $\sim$7B size open-source models.
Notably, the long contexts used in \textsc{In2} training are all synthesized from short segments. 
These improvements suggest that the long-context capabilities learned from the synthesized data can be successfully applied to real-world tasks.

\paragraph{\textsc{FilM-7B} maintains the performance on short-context tasks.}
Figure~\ref{fig:short} illustrates the performances of \textsc{FilM-7B} and the vanilla backbone model on short-context tasks.
It reveals that the overall performances on short-context tasks are almost comparable with minor variances. 
These results confirm that \textsc{FilM-7B} does not compromise the short-context capabilities of the backbone model.

\subsection{Training Strategy Analysis}\label{sec:training_strategy}
Experimental results in Section~\ref{sec:main_results} demonstrate the feasibility of \textsc{In2} training.
We aim to explore further into enhancing the effectiveness and efficiency of \textsc{In2} training, particularly from the perspective of training strategies.
We are specifically interested in investigating the impact of the following two training strategies: applying the sliding window and adjusting the position encoding.
Considering the high cost of training, the following experiments use 20\% of all training examples.

\paragraph{Models using sliding windows cannot effectively capture the long distance information.}
Our experiments involving Mistral models, as shown in Figure~\ref{fig:probing_mistral}, reveal that the performance of Mistral-7B-Instruct-v0.1 is awful when the information is positioned at a long distance. 
It's worth noting that Mistral-7B-Instruct-v0.1 employs the sliding window strategy while Mistral-7B-Instruct-v0.2 does not.
Consequently, we are interested in determining whether our \textsc{In2} training can still alleviate the lost-in-the-middle problem under the sliding window strategy.
We conduct the following two experiments with a 4K sliding window during training:
\begin{itemize}
    \item \textbf{Apply the sliding window in both pre-training and \textsc{In2} training.} We take the Mistral-7B-Instruct-v0.1 as the backbone model and conduct \textsc{In2} training with the same window size (4K).
    \item \textbf{Apply the sliding window only during the \textsc{In2} training.} We take the Mistral-7B-Instruct-v0.2 as the backbone model and additionally apply a 4K sliding window during \textsc{In2} training.
\end{itemize}
Figure~\ref{fig:sw} illustrates the performances of models with sliding windows.
It shows that in both two settings with sliding windows, the performances drop dramatically when the distance between the retrieval question and information is longer than the sliding window size.
It reveals that the sliding window strategy greatly hurts the long-context capability of models.

\paragraph{Training with higher information intensity requires a larger RoPE base $\theta$.}
The training stage in Section~\ref{sec:in2} follows the RoPE settings configured for the backbone model. 
Previous studies on context extension suggest that training with an extended context length necessitates a larger RoPE base $\theta$~\citep{roziere2023code, xiong2023effective, cai2024internlm2}.
In the case of our \textsc{In2} training, the context length remains unchanged, but the information intensity is significantly increased. 
As a result, we are interested in exploring whether the RoPE settings should also be adjusted to further enhance the \textsc{In2} training.
Table~\ref{tab:rope} shows the results with increasing the RoPE base $\theta$ from $1.0 \times 10^6$ to $1.0 \times 10^8$.
It shows that increasing the default RoPE base $\theta$ of the backbone model leads to better performances on \textsc{VaL} Probing.
We suggest to use a 10 times of the default RoPE base $\theta$ to conduct \textsc{In2} training.

\section{Related Work}\label{sec:related}

\paragraph{Long-context LLMs.}
Recent research has significantly contributed to the exploration of training large models with extended context windows~\citep{jiang2023mistral, du2022glm, li2023long, mosaicml2023introducing, together2023, xiong2023effective, song2023hierarchical, tworkowski2024focused, ai2024yi, cai2024internlm2}.
There are primarily two directions in the development of long-context LLMs.
(1) Data engineering, which emphasizes the construction of long-context data for training the LLMs. 
This includes data balancing~\citep{fu2024data}, data order arrangement~\citep{shi2023context}, instruction data collection~\citep{bai2024longalign}, and data quality measurement~\citep{lv2024longwanjuan}. 
Our \textsc{In2} training can be categorized into this field.
(2) Effective and efficient training, which investigates methods to optimize the training of a long-context model.
This encompasses the design of position encoding~\citep{chen2023extending, liu2023scaling, peng2023yarn, ding2024longrope}, batching strategy~\citep{bai2024longalign}, parameter-efficient training~\citep{chen2023longlora}, and the development of new model architectures~\citep{peng2023rwkv, gu2023mamba}.


\paragraph{Long-context evaluations.}
Existing benchmarks for evaluating long-context models can be divided into two categories.
(1) Real-world benchmarks that assess general long-context capabilities (e.g., long-context QA, summarization, and language modeling), such as NarrativeQA~\citep{kovcisky2018narrativeqa}, LongBench~\citep{bai2023longbench}, ZeroSCROLLS~\citep{shaham2023zeroscrolls}, L-Eval~\citep{an2023eval}, Loogle~\citep{li2023loogle}, $\infty$Bench~\citep{zhang2024inftybench}, and a series of work on perplexity evaluation~\citep{beltagy2020longformer, roy2021efficient, press2021train, chen2023extending, liu2023scaling, peng2023yarn, chen2023longlora, ding2024longrope, mohtashami2024random}.
(2) Probing tasks that provide a more concise reflection of the long-context utilization across different context lengths and positions. These include Needle-in-the-Haystack, passkey retrieval~\citep{mohtashami2024random}, synthesized document QA~\citep{liu2024lost}, S3Eval~\citep{lei2024s3eval}, Discovery~\citep{li2024longcontext}, RULER~\citep{hsieh2024ruler}, and the \textsc{VaL} Probing proposed in this study.
Among these probing tasks, our \textsc{VaL} Probing is the first to explicitly incorporate a variety of retrieval patterns.



\section{Conclusion}
This work introduces \textsc{In2} training to overcome the lost-in-the-middle problem.
By applying \textsc{In2} training on the open-source model, our \textsc{FilM-7B} exhibits significant improvements on probing tasks and real-world long-context tasks while does not compromise the short-context performance.

\section*{Acknowledgments}
Shengnan An and Nanning Zheng were supported in part by NSFC under grant No. 62088102.
Thank you to arXiv for use of its open access interoperability.

\bibliography{colm2024_conference}
\bibliographystyle{colm2024_conference}


\newpage
\appendix

This is the Appendix of the paper: \textit{Make Your LLM Fully Utilize the Context}.

\section{Data Filtering Strategy}\label{sec:ap_filter}To avoid data contamination for the evaluation stage in Section~\ref{sec:experiments}, we apply a pre-filtering strategy during sampling the raw texts for constructing the dataset of \textsc{In2} training.
Specifically, during sampling $\mathcal{C}_{i}$ for generating data, if the sampled $\mathcal{C}_{i}$ has a 10-gram overlap with any example in all of our evaluation data (including probing tasks, real-world tasks and short-context tasks), it will not be used for neither generating question-answer pairs nor serving as the random segments $[r_{j}]$.

\section{Training Examples for \textsc{In2} Training}\label{sec:ap_examples}


\begin{exmp}{Fine-Grained Information Awareness (1)}{data_fine_1}
    \small

    \textbf{Segment:}
    
    Last summer I took a cross-country road trip with a group of people driving different types of electric vehicles.
    It was an amazing experience, and our trip, our adventures, and our misadventures were portrayed in the documentary Kick Gas.
    That experience gave me the idea to take my own road trip to experience the freedom of traveling alone on an electric motorcycle.
    My goals were simple: ride the great motorcycle roads of America, meet other electric motorcycle riders, and visit friends  ...

    \textbf{Question:}

    What were the primary goals of the individual's solo road trip on an electric motorcycle?
    
    \textbf{Answer:}

    The individual's primary goals for the solo road trip on an electric motorcycle were to ride the great motorcycle roads of America, meet other electric motorcycle riders, and visit friends.

\end{exmp}

\begin{exmp}{Fine-Grained Information Awareness (2)}{data_fine_2}
    \small

    \textbf{Segment:}
    
    Another addition to the kernel is the ar5523 driver, which was started over five years ago for the Atheros USB chipset of the same name. 
    Extensions for supporting more chips and WLAN adapters were added to a number of other drivers; the brcmsmac WLAN driver, for example, now supports the BCM43224 Broadcom chip, while the rt2800usb RaLink driver supports the Sweex LW323 USB WLAN adapter ...

    \textbf{Question:}

    What new addition to the kernel supports the BCM43224 Broadcom chip?
    
    \textbf{Answer:}

    The new addition to the kernel that supports the BCM43224 Broadcom chip is the brcmsmac WLAN driver.

\end{exmp}

\begin{exmp}{Fine-Grained Information Awareness (3)}{data_fine_3}
    \small

    \textbf{Segment:}
    
    The Arlington County Board plans to vote Saturday afternoon on giving Amazon \$23 million and other incentives to build a headquarters campus in Crystal City, but only after hearing scores of northern Virginia residents and advocates testify for or against the project.
    The five-member board is expected to support the plan, which was announced amid much hoopla on Nov. 13. The proposed county incentives are part of an agreement in which Amazon would occupy significant office space and bring at least 25,000 high-paying jobs to Arlington in coming years.
    Opponents hope to postpone the vote until after additional public hearings, where they want representatives of the online retail giant to answer questions directly from anyone in the community.

    \textbf{Question:}

    What is the Arlington County Board's plan regarding Amazon's proposed headquarters in Crystal City?
    
    \textbf{Answer:}

    The Arlington County Board is planning to vote on providing Amazon with \$23 million and other incentives to construct a headquarters campus in Crystal City. However, this vote will only take place after listening to the opinions of numerous residents and advocates from northern Virginia. The board, which consists of five members, is anticipated to endorse this plan. The proposal includes Amazon occupying a significant amount of office space and introducing at least 25,000 high-paying jobs to Arlington in the forthcoming years.

\end{exmp}

\begin{exmp}{Integration and Reasoning of Information (1)}{data_reason_1}
    \small

    \textbf{Segment 1:}
    
    Cloud Cruiser's utilization analytics application enables users to manage and optimize public, private and hybrid cloud usage and spending ...

    \textbf{Segment 2:}

    This reduces the risk of organizations investing too much or too little in IT, eliminates unused capacity and frees up valuable IT resources for new value-adding projects ...

    \textbf{Segment 3:}
    
    As a Cloud Cruiser customer, we have seen firsthand the value that Cloud Cruiser's technology creates by enabling HPE Flexible Capacity to meter and bill for usage of on-premise IT infrastructure in a pay-as-you-go model ...

    \textbf{Question:}

    What benefits does the acquisition of Cloud Cruiser bring to HPE and its Flexible Capacity business?
    
    \textbf{Answer:}

    The acquisition of Cloud Cruiser enriches HPE's Flexible Capacity business by providing a utilization analytics application that allows users to manage and optimize their cloud usage and spending. Cloud Cruiser's software also facilitates the metering and billing for customers' IT consumption. This enables HPE to offer its customers an on-premise physical IT infrastructure at the cost of cloud services, effectively a rental agreement. Moreover, it reduces the risk of organizations either under or over investing in IT, eliminates unused capacity, and frees up IT resources for new projects ...

\end{exmp}

\begin{exmp}{Integration and Reasoning of Information (2)}{data_reason_2}
    \small

    \textbf{Segment 1:}
    
    Julie Annette Wasden, 45, of 38 Roles St., Haines City, was charged with retail theft at Beall's at 5998 Cypress Gardens Blvd ...

    \textbf{Segment 2:}

    Katherine D. Collazo Santos, 18, of 108 Hilltop Drive, Winter Haven, was charged with retail theft at Macy's at 700 Third St., S.W. ...

    \textbf{Question:}

    Who was charged with retail theft at Beall's and who was charged with the same crime at Macy's?
    
    \textbf{Answer:}

    Julie Annette Wasden and Katherine D. Collazo Santos

\end{exmp}

\begin{exmp}{Integration and Reasoning of Information (3)}{data_reason_3}
    \small

    \textbf{Segment 1:}
    
    A court presided by Magistrate Joseph Mifsud explained in its judgement of 17 August, 2016 in The Police v Eebis Getu, that the charges of a false criminal report and calumnious accusations may be factually similar, however, from a legal point of view they are two separate and distinct charges.
    The accused, Eebis Getu was charged with having filed a false police report and with not obeying a legitimate order. She admitted these charges ...

    \textbf{Segment 2:}

    In passing judgement the court took into consideration that the accused admitted to the charges immediately and also that she did what she did in order to be with her husband in Malta.
    Magistrate Mifsud referred to what Pope Francis said last June, where today\u2019s information technology brings suffering of others instantly, but we also become immune to tragedies and sufferings ...

    \textbf{Question:}

    What were the two charges Eebis Getu admitted to, and what was her reason for committing these actions according to the court's judgement?
    
    \textbf{Answer:}

    Filing a false police report and not obeying a legitimate order; to be with her husband in Malta.

\end{exmp}

\section{Performance on Needle-in-the-Haystack}\label{sec:ap_needle}

\begin{figure*}[ht]
    \centering
    \includegraphics[width=.8\textwidth]{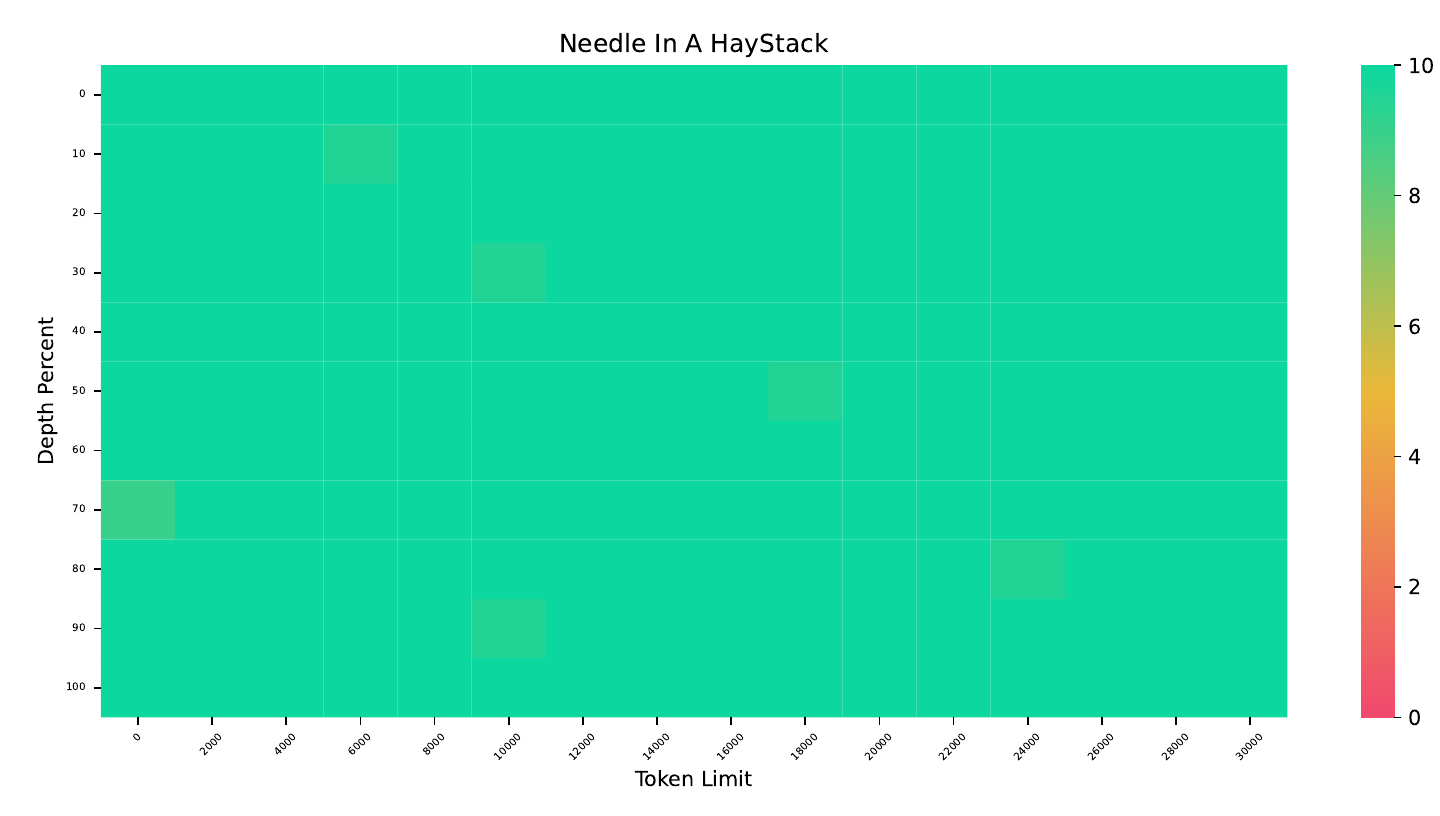}
    \caption{
    Performances of \textsc{FilM-7B} on Needle-in-the-Haystack.
    }
    \label{fig:needle}
\end{figure*}

Figure~\ref{fig:needle} shows the performance of \textsc{FilM-7B} on Needle-in-the-Haystack.
It shows that \textsc{FilM-7B} has achieved near-perfect performance on Needle-in-the-Haystack within its 32K context window.

\section{Prompts For Data Generation and Training}\label{sec:ap_prompts}

\begin{exmp}{Prompt For Equation~\ref{equ:fine}}{prompt_fine}
    \small
    
    Generate one question and the answer from the given context.
    The question should be highly specific to the information provided in the context.
    It should not be a general question that suits any context.\\
    Rules to follow when generate the question:\\
    1. The question should be fully answerable from information present in given context.\\
    2. Make sure the question is clear and unambiguous.\\
    3. Phrases like 'based on the provided context', 'according to the context', etc, are not allowed to appear in the question.\\
    Rules to follow when generate the answer:\\
    1. The answer must use the information provided in the context.\\
    2. Do not just copy words from the context. Answer the question in your own words.\\

    \#\#\# Context \#\#\#:\\
    $s_{i}$\\

    \#\#\# Question \#\#\#:\\
    \{completion\}

\end{exmp}

\begin{exmp}{Prompt For Equation~\ref{equ:reason}}{prompt_reason}
    \small
    
    Generate one question and the answer from the given context.
    The context contains several pieces.
    Answering the question should require the reader to make multiple logical connections or inferences using **at least two pieces**.\\
    Rules to follow when generate the question:\\
    1. The question should be fully answerable from information present in given context.\\
    2. Make sure the question is clear and unambiguous.\\
    3. Phrases like 'based on the provided context', 'according to the context', etc, are not allowed to appear in the question.\\
    Rules to follow when generate the answer:\\
    1. The answer must use the information provided in the context.\\
    2. Do not just copy words from the context. Answer the question in your own words.\\

    \#\#\# Context \#\#\#:\\
    \# Piece 1: $s_{i}^{1}$\\
    \# Piece 2: $s_{i}^{2}$\\
    ...\\

    \#\#\# Question \#\#\#:\\
    \{completion\}

\end{exmp}

\begin{exmp}{Training Template}{prompt_training}
    \small

    \textbf{Input:}

    [INST] Below is a context and an instruction.
    Based on the information provided in the context, write a response for the instruction.\\

    \#\#\# Context:\\
    $\mathcal{L}_{i}$\\

    \#\#\# Instruction:\\
    $q_{i}$ [/INST]\\

    \textbf{Output:}\\
    $a_{i}$

\end{exmp}


\end{document}